\PassOptionsToPackage{table}{xcolor}
\documentclass{article}
\usepackage[T1]{fontenc}
\usepackage{pointtool_preprint,times}

\usepackage{amsmath}
\usepackage{amssymb}
\usepackage{booktabs}
\usepackage{graphicx}
\usepackage{multirow}
\usepackage{tabularx}
\newcolumntype{C}{>{\centering\arraybackslash}X}
\usepackage{capt-of}
\usepackage{xcolor}
\usepackage{pifont}
\usepackage{hyperref}
\hypersetup{
  hidelinks,
  pdftitle={Unleashing the Value of Raw Point Clouds for 3D Understanding with LLMs},
  pdfauthor={Tailai Chen; Jiawei Yu; Yiming Cheng; Ping Nie; Yongyong Chen; Tao Zhou; Zheng Zhu; Ming Li},
  pdfsubject={3D understanding with large language models}
}
\usepackage{enumitem} 
\usepackage{float}
\usepackage{subcaption}

\newcommand{\method}{\textsc{PointTool}}
\definecolor{ours}{RGB}{232,240,252}

\newcommand{\best}[1]{\textbf{#1}}
\newcommand{\second}[1]{\underline{#1}}

\title{Unleashing the Value of Raw Point Clouds for 3D Understanding with LLMs}

\author{
\begin{tabular}{c}
Tailai Chen$^{1,\ast}$, Jiawei Yu$^{2,\ast}$, Yiming Cheng$^3$, Ping Nie$^4$ \\
Yongyong Chen$^5$, Tao Zhou$^6$, Zheng Zhu$^7$, Ming Li$^8$ \\[0.35em]
{\normalfont\scriptsize $^1$Institute of Automation, Chinese Academy of Sciences \quad $^2$National University of Defense Technology} \\
{\normalfont\scriptsize $^3$Tsinghua University \quad $^4$University of Waterloo \quad $^5$Harbin Institute of Technology} \\
{\normalfont\scriptsize $^6$Nanjing University of Science and Technology \quad $^7$GigaAI} \\
{\normalfont\scriptsize $^8$The Chinese University of Hong Kong, Shenzhen \quad $^\ast$Equal contribution}
\end{tabular}
}

\begin{document}

\maketitle

\begin{abstract}

Standard 3D point cloud language models typically rely on pretrained point encoders to extract 3D features and project them into representations processable by large language models (LLMs). However, these encoders are often originally optimized for pre-defined tasks, resulting in a \textit{representation bottleneck} that may lead to a substantial mismatch with downstream 3D understanding. Moreover, existing models largely follow a \textit{single pass closed-book} generation, limiting their ability to actively inspect task-relevant 3D regions or retrieve necessary external knowledge. In this work, we present \textsc{PointTool}, an encoder-free and tool-augmented framework for 3D reasoning. \textsc{PointTool} introduces Alignment-Free Patch Tokenization (AFPT) to locally aggregate raw point clouds while preserving fine-grained geometric structure, and Decoupled Point-Text LoRA (DPL) to facilitate global aggregation and 3D knowledge internalization within a frozen LLM. Building upon these designs, Program-Supervised Tool Reasoning (PSTR) enables the model to explicitly operate on raw 3D data and external resources through executable tools, supporting interactive 3D reasoning grounded in spatial evidence. We further introduce PointCloud-TR, a benchmark comprising 13K 3D objects, 39.8K questions, and 35.6K tool trajectories to facilitate training and evaluation of complex 3D reasoning. Experiments demonstrate that \textsc{PointTool} achieves strong performance on both standard and complex 3D understanding tasks, validating the effectiveness of the proposed framework.
\end{abstract}
\begin{figure}[h]
    \centering
    \includegraphics[width=\linewidth]{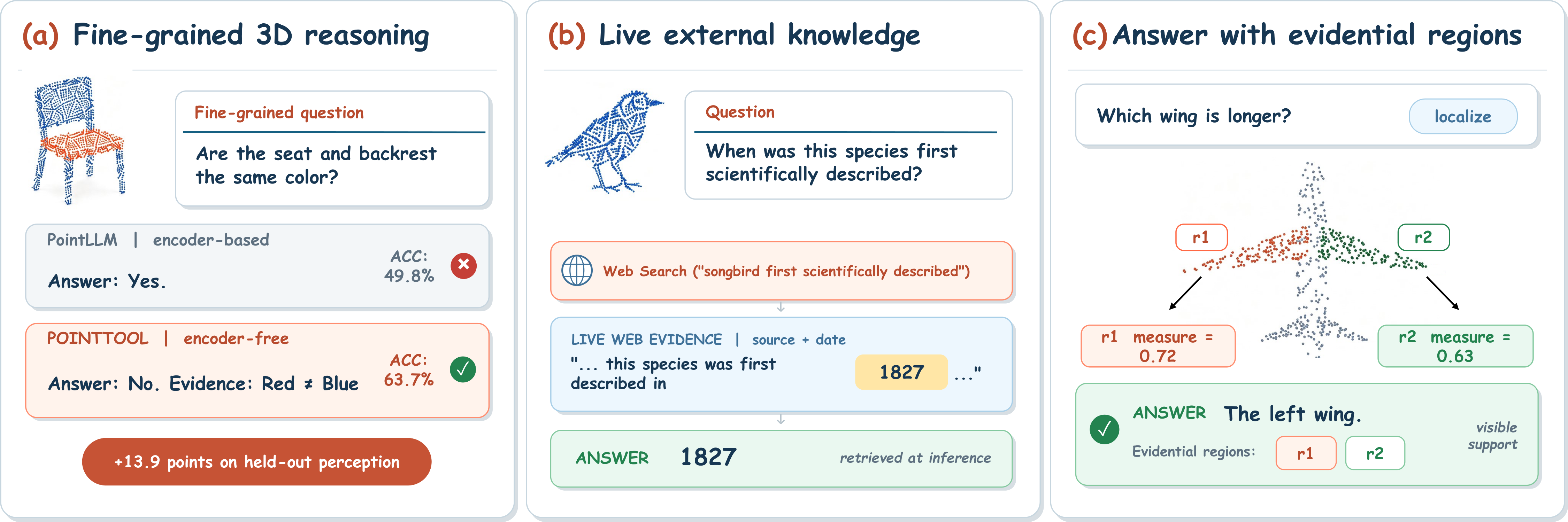}
    \caption{\textsc{PointTool} achieves fine-grained reasoning in an encoder-free manner and leverages tool calls to access external knowledge and task-relevant evidence from raw data for 3D understanding.}
    \label{fig:motivation}
\end{figure}
\section{Introduction}
\label{sec:intro}

3D visual understanding extracts semantic, geometric, and spatial information from three-dimensional data for tasks such as object classification and detection \citep{pointnet,voxelnet}, language-based object localisation and dense captioning \citep{scanrefer,scan2cap}, and 3D question answering \citep{scanqa}. Multimodal LLMs combine visual perception with linguistic knowledge \citep{wu2023multimodal}; recent 3D LMMs extend this paradigm to point clouds and reconstructed scenes \citep{pointllm,3dllm,ll3da}. Within this line of work, point clouds are a common native 3D input \citep{pointllm,shapellm,minigpt3d}, encoding geometry and, when available, color .

Existing mainstream 3D LMM paradigms predominantly employ pretrained 3D point encoders to extract point cloud features \citep{pointllm,shapellm,minigpt3d}. Representative encoders are often pretrained with self-supervised objectives on point clouds \citep{pointbert,pointmae}. Because these objectives do not directly use language supervision, their outputs are not explicitly optimized for language-conditioned downstream tasks. ULIP and ULIP-2 align point-cloud, image, and language representations during pre-training \citep{ulip,ulip2}, while Point-Bind aligns point clouds with images, language, audio, and video in a shared embedding space \citep{pointbind}. These approaches are effective, but a pretrained encoder limits the information exposed to the LLM and can therefore form a \textbf{representation bottleneck} when a downstream question depends on signals not emphasized by pre-training. 

Recent encoder-free 3D LMMs learn point representations more directly with the LLM, but their tokenizers still introduce substantial structural or representational constraints. ENEL \citep{enel} relies on architecture-specific hierarchical aggregation and propagation, which repeatedly compress and reconstruct point features and may obscure fine-grained local information. SAGE \citep{sage} further discretizes continuous point features through a finite codebook, introducing a quantization bottleneck that may collapse subtle geometric and appearance variations. Moreover, both methods remain closed-book, single-pass generators: they cannot invoke tools to dynamically inspect the underlying point cloud or access external knowledge beyond the LLM's parametric memory.
In Fig.~\ref{fig:motivation}, baseline is prone to errors in fine-grained spatial reasoning (a) and fails to answer knowledge-intensive questions that require information beyond the point cloud (b). 

Motivated by these limitations, we present \textbf{\textsc{PointTool}}, an encoder-free framework that exposes fine-grained point-cloud information directly to the LLM. \textsc{PointTool} introduces Alignment-Free Patch Tokenization (AFPT), which groups raw points into local patches and represents each patch as a continuous, position-aware token. By performing only local aggregation, AFPT preserves fine-grained geometric and color information without architecture-specific hierarchical compression or discrete codebook quantization. We further introduce Decoupled Point-Text LoRA (DPL), which provides separate low-rank adaptation paths for point and text tokens while allowing the LLM to model global interactions across patches. Together, AFPT and DPL establish a direct and information-preserving point-language interface, enabling task-dependent 3D modeling without relying on a pretrained point encoder or a restrictive tokenization pipeline.

To further exploit raw point-cloud information and extend reasoning beyond the LLM's parametric knowledge, \textsc{PointTool} introduces Program-Supervised Tool Reasoning (PSTR), which decomposes complex 3D questions into executable programs for point-cloud inspection, numerical computation, and external knowledge retrieval. Through iterative tool interaction, \textsc{PointTool} dynamically acquires task-relevant information and localizes the point-cloud regions supporting its predictions. To train and evaluate this capability, we construct \textbf{PointCloud-TR}, a benchmark for tool-augmented 3D reasoning comprising 13K 3D objects, 39.8K question-answer pairs, and 35.6K executable tool trajectories across point-cloud perception, external-knowledge, and no-tool reasoning tasks. As illustrated in Fig.~\ref{fig:motivation}, this framework enables fine-grained, knowledge-intensive, and evidence-grounded 3D question answering. Our key contributions are summarized as follows:
\begin{itemize}[leftmargin=*]
    \item We introduce \textsc{PointTool}, an encoder-free, tool-augmented 3D language model that unifies direct point-cloud modeling with interactive tool reasoning.

    \item We develop Alignment-Free Patch Tokenization  and Decoupled Point-Text LoRA. AFPT preserves fine-grained geometric, positional, and color information through continuous local patch tokens, while DPL enables a frozen LLM to adaptively model point and text tokens through modality-specific low-rank branches.

    \item We propose Program-Supervised Tool Reasoning, which teaches the model to decompose complex 3D questions into executable programs and iteratively invoke point-cloud operations, numerical computation, and external knowledge retrieval.

    \item We construct PointCloud-TR, a benchmark for tool-augmented 3D reasoning comprising 13K 3D objects, 39.8K question-answer pairs, and 35.6K executable tool trajectories across point-cloud perception, external-knowledge, and no-tool reasoning tasks, providing a comprehensive testbed for evaluating interactive and evidence-grounded 3D reasoning.
\end{itemize}

\section{Related Work}
\label{sec:related}
\textbf{3D Large Multimodal Models.} 
Early works focused on establishing unified 3D-language alignment\citep{pointllm,pointbert,ulip2}. Point-Bind aligned point clouds with multiple modalities in a shared embedding space \citep{pointbind}, while ShapeLLM built on ReCon with multi-view distillation \citep{shapellm,recon}. MiniGPT-3D exploited 2D visual priors for efficient 3D--LLM alignment \citep{minigpt3d}, and GPT4Point unified object-level understanding and generation \citep{gpt4point}. PGLLM \citep{PGLLM} explores test-time optimization of 3D LMMs by constructing a feature-space neighborhood graph over point clouds. Other systems extended 3D LMMs to scene-level understanding \citep{3dllm,chat3d,scenellm,ll3da}. PerLA\citep{Perla} combines fine local details with coarse global context for fine-grained scene understanding. These methods rely on pretrained feature encoders as the interface to the LLM, making the information available to the LLM dependent on the encoder. 

\textbf{Encoder-Free Vision-Language Models.}
To overcome the representational limitations of pretrained 3D encoders and further reduce computational costs, recent studies have explored encoder-free 3D LMMs. ENEL \citep{enel} showed that encoder-free architectures can match encoder-based methods on multiple 3D understanding tasks.
SAGE \citep{sage} introduced a lightweight 3D tokenizer and semantic alignment rewards to improve LLMs' processing of raw point clouds. Fase3D \citep{fase3d} further employed Fourier transforms to approximate self-attention, reducing both computational and parameter costs. 

\textbf{Tool-Augmented Language Models.}
Tool use provides an effective paradigm for extending the capabilities of language models. Toolformer learned when and how to invoke APIs through self-supervised training over unlabeled text, seeded by a handful of demonstrations for each API \citep{toolformer}, while Gorilla focused on accurate API invocation over large tool collections \citep{gorilla}. ViperGPT and VisProg generate executable programs over visual modules \citep{vipergpt,visprog}, whereas MM-ReAct prompts ChatGPT to call vision experts and integrate their outputs \citep{mmreact}. NS3D \citep{NS3D} casts 3D grounding as hierarchical programs executed by neural modules, exemplifying programmatic reasoning.
In contrast to prior works, we aim to leverage this paradigm to enhance LLMs' interactive 3D understanding.

\section{Motivating Analysis}
\label{sec:motivation}

\begin{figure}[htbp]
    \centering
\includegraphics[width=0.98\linewidth]{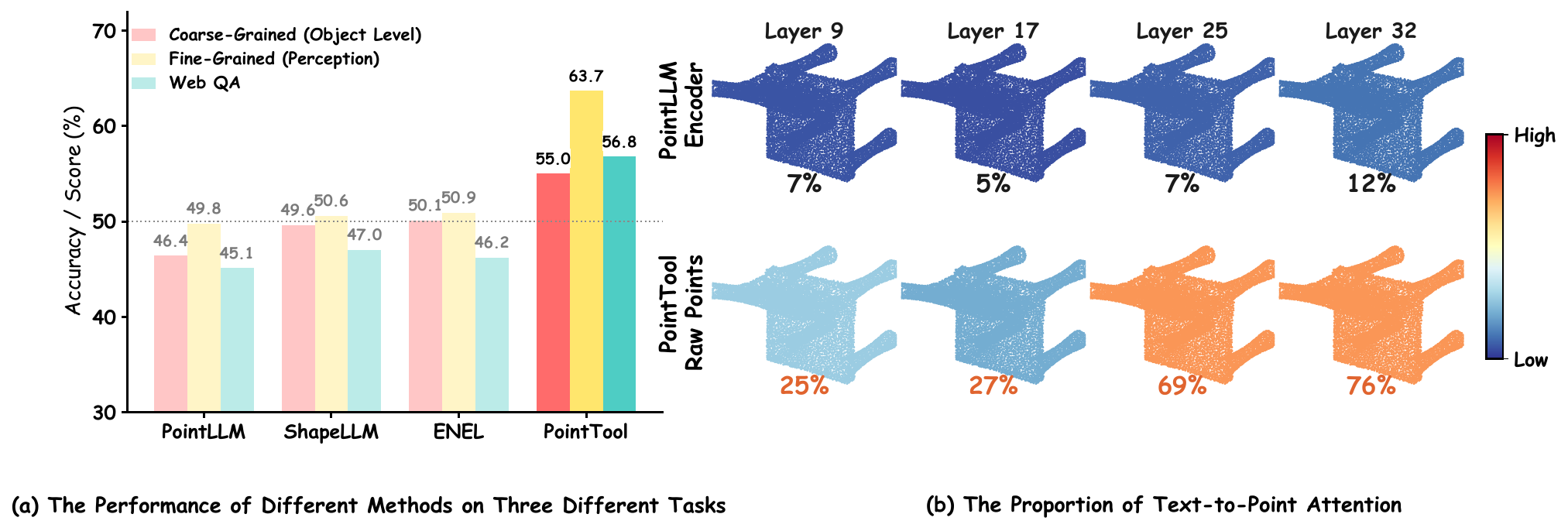}
    \caption{\textbf{Motivating analysis of the 3D representation bottleneck.}
    (a) The performance of different methods on three different tasks.
    (b) The proportion of text-to-point attention.}
    \label{fig:motivation_analysis}
\end{figure}

As shown in Fig.~\ref{fig:motivation_analysis}(a), the pretrained-encoder-based
PointLLM and ShapeLLM and the encoder-free ENEL remain close on fine-grained
perception. In contrast, \method{} improves fine-grained perception while
also strengthening coarse-grained performance. Its gain on Web QA further
shows the value of combining point-cloud evidence with external knowledge.

To examine this difference, we measure text-to-point attention across
LLM layers. In Fig.~\ref{fig:motivation_analysis}(b), PointLLM keeps this
share low after the earliest layers, whereas \method{} increases it above
70\% in the final layer.
\section{Methodology}
\label{sec:method}

\begin{figure}[!b]
    \centering
        \includegraphics[width=1.0\linewidth]{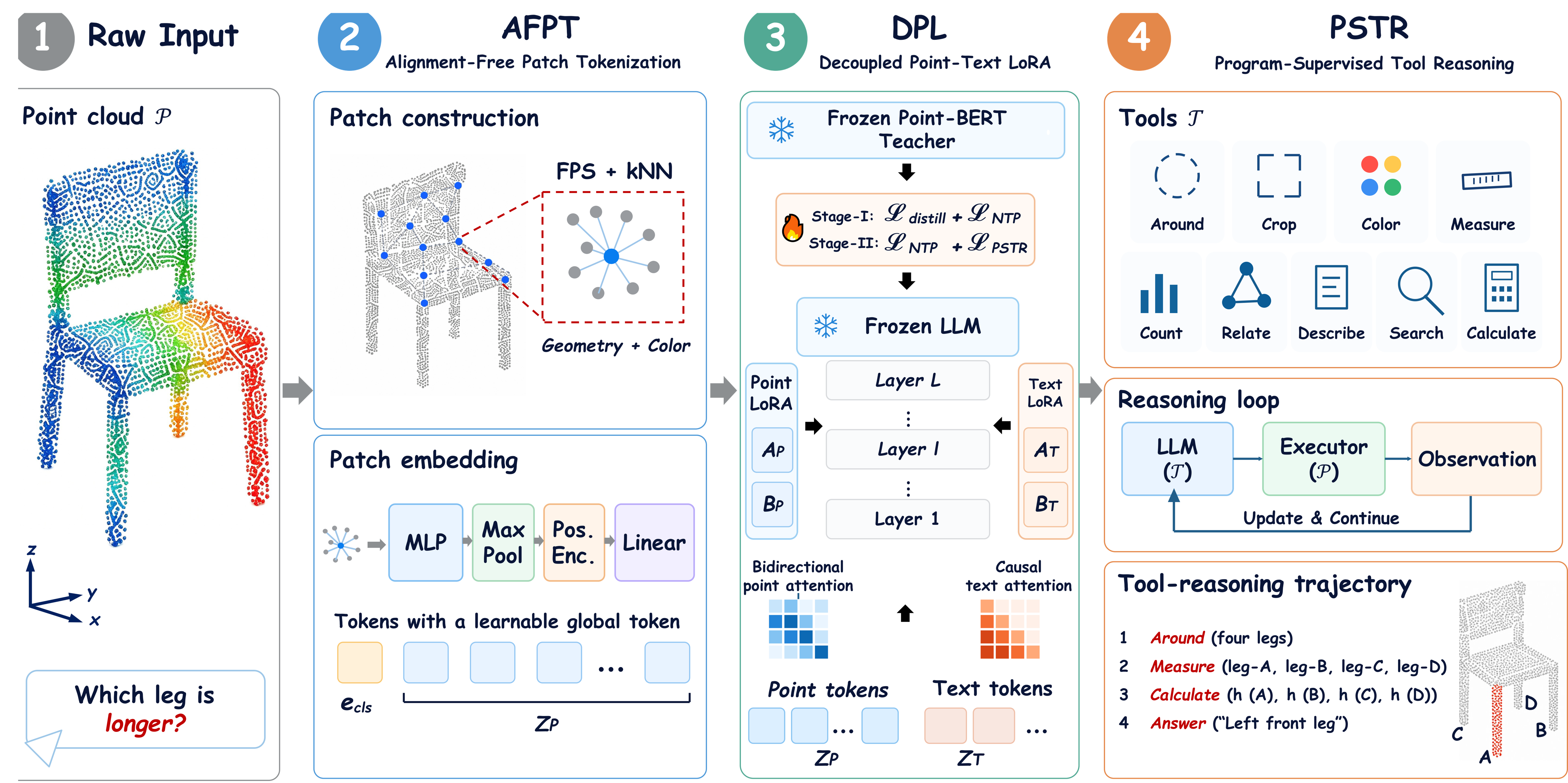}
    \caption{\textbf{Overview of \textsc{PointTool}.} (1) \textbf{AFPT} converts point coordinates and colors into position-aware point tokens \(Z_P\) via farthest point sampling (FPS) and \(k\)NN grouping. (2) \textbf{DPL} adapts the frozen LLM with decoupled Point/Text LoRA and bidirectional point attention, plus feature distillation for 3D structure internalization produced by AFPT. (3) \textbf{PSTR} uses tools on raw point clouds and external resources through an action--observation loop, enabling interactive 3D reasoning grounded in spatial evidence.}
    \label{fig:overview}
\end{figure}
In this section, we present \textsc{PointTool}, an encoder-free and tool-augmented framework designed to fully exploit information inherent in raw point clouds for complex 3D reasoning. The overall architecture of \textsc{PointTool} is illustrated in Fig.~\ref{fig:overview}, which consists of three key components.

\subsection{Alignment-Free Patch Tokenization (AFPT)}
\label{sec:tokenizer}

Unlike conventional pretrained 3D encoders that compress the entire point cloud into a compact feature representation, AFPT performs only local aggregation, preserving geometric details from the raw point cloud before LLM processing. It keeps each patch as an independent point token and defers cross-patch interaction to the language model for downstream task-dependent modeling.

Formally, given a raw point cloud
\(P=\left\{\left(p_n^{xyz},p_n^{rgb}\right)\right\}_{n=1}^{N},\)
where \(p_n^{xyz}, p_n^{rgb}\) denote the 3D coordinates and color attributes of the $n$-th point, respectively. AFPT first employs the iterative \textit{Farthest Point Sampling} (FPS) routine used by PointNet++ \citep{pointnet2} to select $M$ center points:
\(C=\{c_i\}_{i=1}^{M}\). For each center point $c_i$, its $k$ nearest neighbors are retrieved from the raw point cloud to construct a local point patch:
\begin{equation}
\mathcal{N}_i=\operatorname{kNN}(c_i,P;k).
\end{equation}
For each local patch, we represent point coordinates in a center-relative coordinate system while retaining the original color attributes: \(\tilde{p}_{ij}=p_{ij}^{xyz}-c_i\),
where $p_{ij}$ denotes the $j$-th neighbor in $\mathcal{N}_i$.

We then apply a shared point-wise mapping $f_{\theta}$ to each point and aggregate the points within each local patch using a symmetric max-pooling operation:
\begin{equation}
e_i=
\operatorname{MaxPool}_{j}
f_{\theta}(x_{ij}),
\end{equation}
where \(x_{ij}=\left[\tilde{p}_{ij},p_{ij}^{rgb}\right]\) is the input feature of the \(j\)-th neighbor in the \(i\)-th patch. \(e_i\) denotes the feature representation of the \(i\)-th local patch.

To preserve the spatial location of each local patch within the entire object, we further introduce a positional encoding
$\phi(c_i)\in\mathbb{R}^{d_p}$ and obtain the position-aware patch representation:
\(\bar{e}_i=e_i+\phi(c_i)\).

Finally, AFPT introduce a learnable global token $e_{\mathrm{cls}}$ and project the resulting point cloud representations into the hidden dimension of the language model:
\begin{equation}
Z_P
=
E_PW_P
=
\left[
e_{\mathrm{cls}};
\bar{e}_1;
\ldots;
\bar{e}_M
\right]W_P,
\qquad
W_P\in\mathbb{R}^{d_p\times d_{\mathrm{LM}}},
\label{eq:point_tokens}
\end{equation}
where
$E_P$ 
and
$Z_P$ 
denote the pre-projection and projected point-token sequences, respectively. $Z_P$ is then inserted into the LLM input at the position corresponding to the referenced point cloud.

\subsection{Decoupled Point-Text LoRA (DPL)}
\label{sec:lora}

AFPT effectively preserves local geometry in the resulting point tokens. However, enabling the LLM to further internalize these 3D representations still poses  challenges. First, point and text tokens are heterogeneous, requiring effective modality adaptation within the shared  LLM. Second, AFPT defers global aggregation, requiring the LLM to recover global 3D structure via cross-region interactions. Third, a mechanism is needed to enable the LLM to acquire sufficient 3D structural knowledge. To this end, we propose DPL, which integrates three complementary components.

\paragraph{Modality-Specific Adaptation.}

To accommodate the point-text heterogeneity, DPL assigns separate Low-Rank Adaptation (LoRA) branches \citep{lora} to each modality while keeping the LLM frozen. Formally, consider a frozen linear layer with weight matrix:
\(W\in\mathbb{R}^{d_{\mathrm{out}}\times d_{\mathrm{in}}}\).
For the $j$-th token position, let $h_j$ and $h'_j$ denote its input and output hidden states, respectively. The adapted transformation is 
\(h'_j=Wh_j+\Delta W_jh_j\), 
where DPL selects the low-rank update according to the token type:
\begin{equation}
\Delta W_j=
\frac{\alpha}{r}
\begin{cases}
B_PA_P, & j\in S_P,\\[4pt]
B_TA_T, & j\in S_T,
\end{cases}
\label{eq:dpl}
\end{equation}
$S_P$ and $S_T$ denote the sets of point-token and text-token positions, respectively. The corresponding low-rank parameters are 
\(A_P,A_T\in\mathbb{R}^{r\times d_{\mathrm{in}}}, B_P,B_T\in\mathbb{R}^{d_{\mathrm{out}}\times r}\).

\paragraph{Global Aggregation.}
To enable the LLM to model relationships among different 3D regions, we modify the attention mask for point tokens. Let $\Gamma$ denote the attention mask, for any two point-token positions $i,j\in S_P$, we set
\(\Gamma_{ij}=1\), 
allowing bidirectional attention among point tokens such that each local region can access the complete set of point-token representations. In contrast, text tokens retain the standard causal attention constraint and cannot attend to future text positions:
\(\Gamma_{ij}=0\ (j>i,\ i,j\in S_T)\).

\paragraph{Feature Distillation.}

We introduce a Point-BERT \citep{pointbert} as a training-time teacher to further facilitate point LoRA's internalization of 3D structural knowledge. 
Specifically, 
\begin{equation}
\mathcal{L}_{\mathrm{distill}}
=
\frac{1}{|{L}|M}
\sum_{l\in{L}}
\sum_{i=1}^{M}
\left[
1-
\cos
\left(
\hat{f}_{i}^{(l)},
f_{i}^{\mathrm{T}}
\right)
\right],
\label{eq:distill}
\end{equation}
where \(L\) is the set of LLM layers used for distillation. \(\hat{f}_{i}^{(l)}\) denotes the student feature of patch \(i\) at layer \(l\), and \(f_{i}^{\mathrm{T}}\) the corresponding teacher feature. Teacher network remains frozen during training.

\subsection{Program-Supervised Tool Reasoning (PSTR)}
\label{sec:tools}

Although AFPT and DPL proposed in \textsc{PointTool} alleviate the representation bottleneck, fully exploiting the information inherent in raw point clouds, particularly for complex 3D tasks, often requires explicit operations such as local region inspection, geometric measurement, spatial relation comparison, and even external knowledge retrieval. To address this, we propose PSTR, which formulates 3D understanding as an interactive reasoning process.

\paragraph{Tool Space.}
PSTR defines a unified tool set: 
\begin{equation}
\mathcal{T}=
\{
\operatorname{Locate},
\operatorname{Crop},
\operatorname{Color},
\operatorname{Measure},
\operatorname{Count},
\operatorname{Relate},
\operatorname{Describe},
\operatorname{Search},
\operatorname{Calculate}
\},
\end{equation}
Together with a terminating action $\operatorname{ANSWER}$, these tools form the complete action space $\mathcal{A}=\mathcal{T}\cup\{\operatorname{ANSWER}\}$.

\paragraph{Interactive Execution.}
Given the point-token sequence $Z_P$, a question $q$, and the interaction history
\(\mathcal{H}_{t-1}=\{(a_i,o_i)\}_{i=1}^{t-1}\), where \(a_i \in \mathcal{A}\) and \(o_i\) denote the action and observation at step \(i\), respectively. The next action is predicted as  
\(a_t \sim \pi_{\Theta}(a_t \mid Z_P,q,\mathcal{H}_{t-1})\), where \(\pi_{\Theta}\) denotes the policy implemented by the LLM.
For a tool call, the executor applies \(a_t\) to the environment, returning \(o_t=\mathcal{E}(a_t;P,\mathcal{D})\), \(P\) is the raw point cloud and \(\mathcal{D}\) denotes external data sources when required. 

The resulting observation is then appended to the interaction history: 
\(\mathcal{H}_t=\mathcal{H}_{t-1}\cup\{(a_t,o_t)\}\), and fed back to the model as part of the updated context, allowing subsequent actions to be conditioned on the newly acquired evidence. Region-producing tools additionally return persistent region references, enabling selected areas to be reused and progressively refined across multiple operations.The process terminates when the model generates $\operatorname{ANSWER}$, yielding an interaction trajectory ending with the final answer \(y\): 
\begin{equation}
\tau
=
(a_1,o_1,\ldots,a_K,o_K,y),
\qquad
K\leq 6,
\end{equation}
Through this action--observation loop, the model can actively gather task-relevant evidence from the 3D data, enabling it to dynamically adjust subsequent operations according to the returned tool results and thereby benefit downstream 3D understanding tasks.

\paragraph{Program-Supervised Trajectories.}
The above interaction requires the model to learn not only the final answer, but also when to invoke a tool, which operation to perform, and where to perform it. Supervising only the final answer cannot directly constrain these intermediate decisions. We therefore construct complete supervision trajectories from executable reference programs.

For each training question with known scene structure, we first construct a reference program
\(\pi^{*}=(a_1^{*},\ldots,a_K^{*})\),
where each action is executable within the PSTR tool space. The program is then executed on the raw point cloud to obtain the corresponding observations:
\(o_t^{*}=\mathcal{E}(a_t^{*};P,\mathcal{D})\),
yielding the reference trajectory: 
\begin{equation}
\tau^{*}
=
(a_1^{*},o_1^{*},\ldots,a_K^{*},o_K^{*},
\operatorname{ANSWER}(y^{*})).
\label{eq:reference_traj}
\end{equation}
The model is trained on the executed trajectory, learning each tool invocation and the final answer autoregressively conditioned on the preceding observations. Specifically, the training objective is
\begin{equation}
\mathcal{L}_{\mathrm{PSTR}}
=
-\sum_{t=1}^{K}
\log
\pi_{\Theta}
\left(
a_t^{*}
\mid
Z_P,q,\mathcal{H}_{t-1}^{*}
\right)
-
\log
\pi_{\Theta}
\left(
\operatorname{ANSWER}(y^{*})
\mid
Z_P,q,\mathcal{H}_{K}^{*}
\right),
\label{eq:pstr_loss}
\end{equation}
where $\mathcal{H}_{t-1}^{*}$ contains the previously executed actions and their corresponding observations.

\subsection{Training Strategy}
\label{sec:training}

\textsc{PointTool} is trained in two stages. Stage I jointly trains AFPT and DPL to establish and internalize 3D representations within the LLM, while Stage II further incorporates PSTR to learn instruction following and interactive tool reasoning. Throughout both stages, the LLM backbone remains frozen, while the AFPT tokenizer and Point/Text LoRA branches are optimized.

\paragraph{Next-Token Prediction.}
Both stages adopt a unified autoregressive objective. Given a sequence $X=(x_1,\ldots,x_T)$ and a binary mask $m_t\in\{0,1\}$ indicating supervised positions, we define: 
\begin{equation}
\mathcal{L}_{\mathrm{NTP}}
=
-\frac{1}{\sum_{t=1}^{T}m_t}
\sum_{t=1}^{T}
m_t
\log
p_{\Theta}(x_t\mid Z_P,x_{<t}),
\label{eq:ntp}
\end{equation}
For regular instructions, only response tokens are supervised. For PSTR trajectories, the model-generated tool calls and final answer are supervised, so that Eq.~\ref{eq:ntp} becomes the trajectory objective \(\mathcal{L}_{\mathrm{PSTR}}\) in Eq.~\ref{eq:pstr_loss}.

\paragraph{Stage I:} 
Following PointLLM, Stage I uses 660K brief object descriptions from Objaverse \citep{objaverse,pointllm} and trains AFPT and DPL with the feature distillation objective, providing additional structural supervision for 3D representation learning within the LLM:
\begin{equation}
\mathcal{L}_{\mathrm{stage1}}
=
\mathcal{L}_{\mathrm{NTP}}
+
\lambda_{\mathrm{dist}}
\mathcal{L}_{\mathrm{distill}},
\label{eq:stage1}
\end{equation}
where $\lambda_{\mathrm{dist}}=1$. This stage enables AFPT to preserve and expose fine-grained point cloud information, while DPL progressively adapts and organizes these representations within the LLM.

\paragraph{Stage II:} 
At this stage, we remove the Point-BERT teacher and distillation and jointly train on 70K complex instructions from Objaverse \citep{objaverse,pointllm} and the training split of PointCloud-TR.

Instruction data are supervised with Eq.~\ref{eq:ntp} and tool trajectories with Eq.~\ref{eq:pstr_loss}, which is the same objective under the trajectory mask, so the two sources are optimized jointly:
\begin{equation}
\mathcal{L}_{\mathrm{stage2}}
=
\mathbb{E}_{X\sim\mathcal{D}_{\mathrm{inst}}}
\left[
\mathcal{L}_{\mathrm{NTP}}(X)
\right]
+
\mathbb{E}_{\tau\sim\mathcal{D}_{\mathrm{tool}}}
\left[
\mathcal{L}_{\mathrm{PSTR}}(\tau)
\right].
\label{eq:stage2}
\end{equation}
Following the representation-learning-to-instruction-tuning progression used in prior 3D LMMs \citep{pointllm,enel}, this two-stage design first focuses on 3D knowledge internalization, followed by instruction tuning for both instruction following and tool reasoning.

\section{Experiments}
\label{sec:experiments}

\subsection{Experimental Setup}
\label{sec:setup}

\paragraph{Evaluation Benchmarks.}
We conduct experiments on two complementary suites: standard 3D understanding and tool reasoning. Following the protocol established in prior work \citep{pointllm,enel}, the former covers captioning and classification on Objaverse \citep{objaverse,pointllm} and multimodal VQA on 3D MM-Vet \citep{shapellm}. The latter is evaluated on the held-out split of PointCloud-TR Dataset (Section~\ref{sec:data}), comprising Perception QA, Web QA, and No-tool QA.

\paragraph{Implementation Details.}
Following the setup of the previous work \citet{pointllm}, we adopt Vicuna-7B v1.1 as the frozen LLM backbone. For AFPT, each raw point cloud (\(N=8192\)) is sampled into \(M=512\) local patches with \(k=32\) nearest neighbors per patch, projecting the local patch representations to \(d_p=384\) before LLM projection. For DPL, both Point and Text LoRA adapters are configured with rank \(r=64\) and scaling factor \(\alpha=128\), applied to all linear projection matrices in every Transformer layer. Feature distillation is performed at LLM layers \(L=\{8,16\}\) using a frozen Point-BERT teacher.

\begin{figure}[!htbp]
    \centering
\includegraphics[width=1\linewidth]{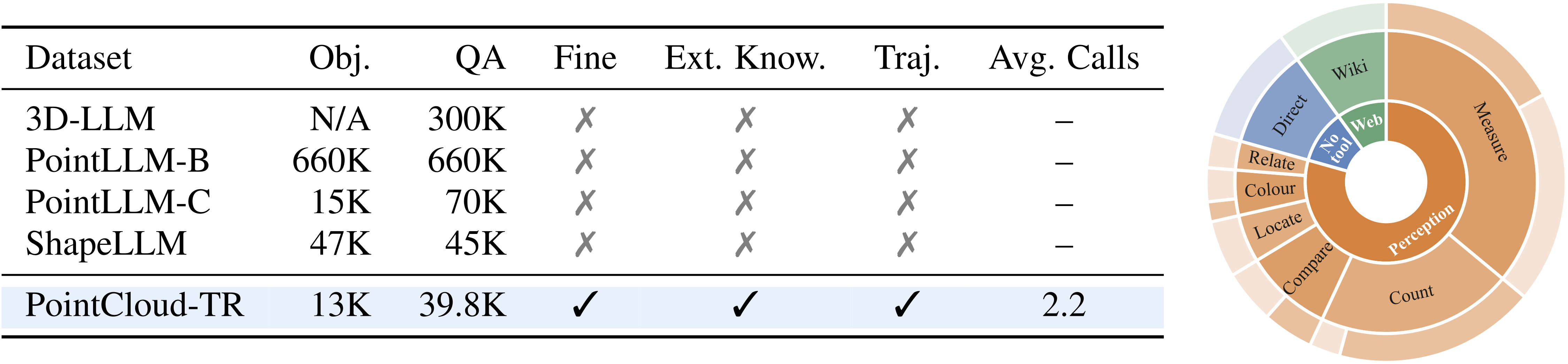}
    \caption{Composition of the Proposed PointCloud-TR Dataset}
    \label{fig:data}
\end{figure}
\subsection{PointCloud-TR Dataset for Tool Reasoning}
\label{sec:data}
To support the training and evaluation of tool-mediated 3D reasoning for challenging tasks, we construct PointCloud-TR, a benchmark comprising diverse types of 3D reasoning data (Fig.~\ref{fig:data}). It contains approximately 13K 3D objects and 39.8K question-answer pairs, accompanied by 35.6K executable tool trajectories with an average of 2.2 tool calls per trajectory.

Existing 3D datasets such as 3D-LLM~\citep{3dllm}, PointLLM~\citep{pointllm}, and ShapeLLM~\citep{shapellm} do not jointly provide fine-grained part queries, retrieval-grounded external-knowledge questions, and executable reasoning trajectories. PointCloud-TR is organized into three key task categories: \textbf{Perception}, which covers geometric and spatial queries, including Measurement (14.3K), Count (8.3K), Comparison (3.7K), Localisation (2.0K), color  (1.9K), and Relation (1.2K); \textbf{Web}, which contains 4.0K Wikipedia-grounded questions; and \textbf{No-tool}, which contains 4.2K questions that can be answered directly without tool use. Together, these categories evaluate not only 3D reasoning, but also the ability to acquire task-relevant evidence and determine when explicit tool use is necessary, providing a dedicated benchmark for 3D understanding.
\begin{table}[t]
    \centering
    \small
    \setlength{\tabcolsep}{4pt}
    \renewcommand{\arraystretch}{1.05}
    \caption{Performance on standard 3D understanding benchmarks: Objaverse and 3D MM-Vet. $\dagger$: same training data as baselines; full \method{} additionally trained on PointCloud-TR.}
    \label{tab:main}
    \begin{tabularx}{\linewidth}{lCCCCC}
        \toprule
        & \multicolumn{3}{c}{Captioning} & Cls. & VQA \\
        \cmidrule(lr){2-4}\cmidrule(lr){5-5}\cmidrule(l){6-6}
        Method & GPT-4 & S-BERT & SimCSE & GPT-4 & GPT-4 \\
        \midrule
        \multicolumn{6}{@{}l}{\textit{Pretrained-encoder-based}} \\
        3D-LLM & 33.42 & 44.48 & 43.68 & 45.25 & -- \\
        PointLLM-7B & 44.85 & 47.47 & 48.55 & 53.00 & 41.20 \\
        PointLLM-13B & 48.15 & 47.91 & 49.12 & 54.00 & 46.60 \\
        ShapeLLM-7B & 46.92 & 48.20 & 49.23 & 54.50 & 47.40 \\
        ShapeLLM-13B & 48.94 & 48.52 & 49.98 & 54.00 & \second{53.10} \\
        MiniGPT-3D & 52.49 & 48.73 & 49.26 & 54.50 & 43.60 \\
        \midrule
        \multicolumn{6}{@{}l}{\textit{Encoder-free}} \\
        ENEL-7B & 51.03 & 48.79 & 49.52 & 55.55 & 43.80 \\
        SAGE-7B & 50.98 & 50.11 & 49.70 & 57.11 & 49.53 \\
        \method-7B$^\dagger$ & \second{52.8} & \second{50.3} & \best{50.3} & \second{57.3} & 51.9 \\
        \rowcolor{ours}
        \method-7B & \best{53.6} & \best{50.6} & \second{50.1} & \best{57.9} & \best{53.4} \\
        \bottomrule
    \end{tabularx}
\end{table}

\subsection{Performance on Standard 3D Understanding Benchmarks}
\label{sec:main}

Table~\ref{tab:main} reports performance on standard 3D understanding benchmarks. While prior encoder-free methods largely match encoder-based counterparts, our \method{} leads four of the five reported metrics and remains close on SimCSE, supporting the benefit of internalizing raw point clouds in the LLM. Specifically, it improves over the strongest prior results by \(1.1\), \(0.8\), and \(0.3\) points on GPT-4 captioning, classification, and VQA, respectively. Notably, the strongest competing VQA result comes from 13B ShapeLLM, whereas ours uses a standard 7B LLM.

We further compare two training variants of \method{}. The $\dagger$
variant is already competitive when trained with the same data as the
competing baselines, while incorporating PointCloud-TR consistently
improves GPT-4 captioning ($+0.8$), Sentence-BERT ($+0.3$), classification
($+0.6$), and VQA ($+1.5$), with only a marginal change in SimCSE ($-0.2$).
These results show that the additional supervision provided by
PointCloud-TR, particularly its substantial collection of executable
tool trajectories, complements instruction tuning without
compromising standard 3D understanding.

\subsection{Performance on the Proposed PointCloud-TR Benchmark}
\label{sec:toolqa}
\begin{table}[b]
    \centering
    \small
    \setlength{\tabcolsep}{4pt}
    \caption{Performance on PointCloud-TR. Overall$_{\rm P+W}$ is the
    question-weighted Perception--Web average; Routing covers all three
    routes, including No-tool.}
    \label{tab:toolqa}
    \begin{tabularx}{\linewidth}{lCCCCC}
        \toprule
        & \multicolumn{3}{c}{Answer $\uparrow$} & & \\
        \cmidrule(lr){2-4}
        Model & Perception & Web & Overall$_{\rm P+W}$ & Routing $\uparrow$ & Execution $\uparrow$ \\
        \midrule
        \multicolumn{6}{@{}l}{\textit{One-pass setting}} \\
        PointLLM-7B         & 49.8 & 45.1 & 49.5 & -- & -- \\
        ShapeLLM-7B         & 50.6 & 47.0 & 50.4 & -- & -- \\
        ENEL-7B                 & 50.9 & 46.2 & 50.6 & -- & -- \\
        \method-7B                           & 51.2 & 47.5 & 50.9 & -- & -- \\
        \method-7B, Answer-only              & 54.6 & 51.2 & 54.4 & -- & -- \\
        \midrule
        \multicolumn{6}{@{}l}{\textit{Tool-reasoning setting}} \\
        PointLLM-7B + PSTR                   & \second{57.6} & \second{52.9} & \second{57.3} & \second{80.1} & \second{89.3} \\
        \rowcolor{ours}
        \method-7B + PSTR                    & \best{63.7} & \best{56.8} & \best{63.2} & \best{83.2} & \best{90.1} \\
        \bottomrule
    \end{tabularx}
\end{table}

We evaluate \method{} and representative baselines on the proposed PointCloud-TR benchmark in Table~\ref{tab:toolqa}. In the one-pass setting, existing methods and \method{} achieve only around 50 Overall$_{\rm P+W}$ accuracy, highlighting the difficulty of the benchmark without explicit tool interaction. In contrast, enabling PSTR improves \method{}'s Overall$_{\rm P+W}$ accuracy from 50.9 to 63.2, with substantial gains on both perception and external-knowledge questions, demonstrating the significant benefit of interactive tools for 3D understanding.

To isolate the source of these gains, the answer-only variant, trained on the same questions but without tool trajectories, reaches only 54.4 Overall$_{\rm P+W}$, far below \method{} with PSTR (63.2). Moreover, PointLLM-7B + PSTR reaches 57.3 Overall$_{\rm P+W}$, while \method{} + PSTR achieves 63.2, further demonstrating the effectiveness of our encoder-free design given the same tool space and program supervision.

\subsection{Ablation Study}
\label{sec:ablation}

\paragraph{Component Ablation.}Table~\ref{tab:ablation} presents the ablation results of the proposed components. Replacing the pretrained Point-BERT encoder(\textit{Original}) with AFPT consistently improves all four metrics, demonstrating the benefits of the encoder-free design. Adding DPL without distillation yields a further $0.5$--$0.6$ point improvement, validating the effectiveness of modality-specific adaptation and in-LLM global aggregation. Finally, feature distillation brings an additional $0.4$--$0.6$ point gain across all metrics, supporting its complementary role in 3D structural representation learning.
\begin{table}[!htbp]
    \centering
    \small
    \setlength{\tabcolsep}{4pt}
    \renewcommand{\arraystretch}{1.05}
    \caption{The ablation results on the proposed components from \method{}.}
    \label{tab:ablation}
    \begin{tabularx}{\linewidth}{lCCCC}
        \toprule
        & \multicolumn{2}{c}{Captioning} & Cls. & VQA \\
        \cmidrule(lr){2-3}\cmidrule(lr){4-4}\cmidrule(l){5-5}
        Component & GPT-4 & S-BERT & GPT-4 & GPT-4 \\
        \midrule
        Original & 50.9 & 48.8 & 55.5 & 49.6 \\
        \quad + AFPT & 51.7 & 49.4 & 56.2 & 50.7 \\
        \quad + DPL, w/o FD & \second{52.2} & \second{49.9} & \second{56.7} & \second{51.3} \\
        \rowcolor{ours}
        \quad + FD (\method-7B$^\dagger$) & \best{52.8} & \best{50.3} & \best{57.3} & \best{51.9} \\
        \bottomrule
    \end{tabularx}
\end{table}
\paragraph{Analysis of Interactive Tool Reasoning.}
\label{sec:tool_analysis}

Beyond overall performance, we analyze the effect of multi-step tool
interaction in Table~\ref{tab:trajectory_analysis}. Compared with direct
answering, predicted tool execution consistently improves accuracy across
different trajectory lengths, with gains of $15.3$, $8.3$, and $4.4$ points
for 2-, 3-, and 5-call trajectories, respectively. The performance gap to
oracle execution also increases for longer trajectories, indicating that
errors accumulate as the interaction becomes more involved. 
Table~\ref{tab:routing_analysis} further shows that PSTR learns to select
appropriate initial actions, achieving $83.9\%$, $77.0\%$, and $80.5\%$
accuracy for Perception, Web, and No-tool questions, respectively. These
results demonstrate that PSTR supports both multi-step tool interaction and
adaptive tool selection.
\begin{table}[htbp]
    \centering
    \begin{minipage}[t]{0.48\linewidth}
        \centering
        \small
        \setlength{\tabcolsep}{4pt}
        \renewcommand{\arraystretch}{1.05}
        \captionof{table}{Multi-step reasoning analysis.}
        \label{tab:trajectory_analysis}
        \begin{tabular*}{\linewidth}{@{\extracolsep{\fill}}lccc@{}}
            \toprule
            \multirow{2}{*}{Calls}
            & \multicolumn{3}{c}{Answer Accuracy (\%)}\\
            \cmidrule(lr){2-4}
            & No tools & Reference & Predicted\\
            \midrule
            2 Calls & 51.9 & 82.1 & 67.2\\
            3 Calls & 50.6 & 77.5 & 58.9\\
            5 Calls & 49.0 & 72.8 & 53.4\\
            \bottomrule
        \end{tabular*}
    \end{minipage}
    \hfill
    \begin{minipage}[t]{0.48\linewidth}
        \centering
        \small
        \setlength{\tabcolsep}{4pt}
        \renewcommand{\arraystretch}{1.05}
        \captionof{table}{Tool routing behavior.}
        \label{tab:routing_analysis}
        \begin{tabular*}{\linewidth}{@{\extracolsep{\fill}}lccc@{}}
            \toprule
            \multirow{2}{*}{Question Type}
            & \multicolumn{3}{c}{First Action (\%)}\\
            \cmidrule(lr){2-4}
            & Perception & Web & Direct\\
            \midrule
            Perception & 83.9 & 4.3 & 11.8\\
            Web & 8.6 & 77.0 & 14.4\\
            No Tool & 13.1 & 6.4 & 80.5\\
            \bottomrule
        \end{tabular*}
    \end{minipage}
\end{table}

\vspace{-1.5em}
\section{Conclusion}
\label{sec:conclusion}
In this work, we present \textsc{PointTool}, an encoder-free, tool-augmented framework for 3D reasoning. AFPT and DPL directly connect raw point clouds to an LLM, preserving fine-grained 3D information while enabling its internalization. PSTR adds interactive reasoning via executable point-cloud operations and external resources, letting the model actively gather task-relevant evidence. We also build PointCloud-TR, a benchmark with executable trajectories for tool-augmented 3D understanding. Experiments show \textsc{PointTool} achieves superior performance on most metrics across standard benchmarks and challenging PointCloud-TR tasks.

\subsection*{AI Use Statement}
Generative AI tools were used to assist with language and presentation,
manuscript formatting, and checking internal consistency and bibliographic
metadata. A large language model was also used inside the data pipeline to
generate candidate web questions from retrieved passages; each candidate is
retained only if it passes the automatic quote and
name-leakage checks. GPT-4 serves as the automatic judge for captioning,
classification and visual question answering, following the protocol of the
benchmarks we report on. All experiments, results and claims are the authors'
own; the authors reviewed every suggested change and take responsibility for
the final content.

\subsection*{Ethics Statement}
This work uses publicly released research datasets, drawing objects from
Objaverse and ShapeNet-Part \citep{objaverse,shapenetpart}. The generated
questions concern object geometry, appearance, and category rather than
personal or identifying information. External evidence
is retrieved from Wikipedia and stored with its article title and revision
date, which both credits the source and keeps the evaluation reproducible.
The model inherits the biases of its frozen language backbone and
of the retrieved text, and the tools return measurements of the input point
cloud rather than assertions about people.

\subsection*{Reproducibility Statement}
The experiments use fixed tool definitions, a deterministic executor, and
stored Wikipedia snapshots. Evaluation follows the public protocols of \citet{pointllm} and
\citet{shapellm}, and web questions are scored against the stored Wikipedia
snapshots so that the numbers do not depend on when the benchmark is run.

\bibliography{references}
\bibliographystyle{pointtool_preprint}
\end{document}